\documentclass[conference]{IEEEtran}

\ifCLASSINFOpdf
   \usepackage[pdftex]{graphicx}
   \graphicspath{{./},{./plots/}}
\else
\fi

\usepackage[cmex10]{amsmath}
\usepackage{amsthm}
\usepackage{float}
\usepackage{caption}
\usepackage[font=footnotesize]{subfig}

\usepackage{mathptmx}

\usepackage{xcolor}
\usepackage{cite}
\begin{document}
%
\title{Exploring Transfer Function Nonlinearity in Echo State Networks}

\author{\IEEEauthorblockN{Alireza Goudarzi}
\IEEEauthorblockA{Department of Computer Science\\
University of New Mexico\\
Albuquerque, NM 87131\\
Email: alirezag@cs.unm.edu}
\and
\IEEEauthorblockN{Alireza Shabani}
\IEEEauthorblockA{Google Inc.\\
 Venice, CA 90291\\
Email: shabani@google.com}
\and
\IEEEauthorblockN{Darko Stefanovic}
\IEEEauthorblockA{Department of Computer Science and \\ 
Center for Biomedical Engineering\\
University of New Mexico\\
Albuquerque, NM 87131\\
Email: darko@cs.unm.edu}}

\maketitle

\begin{abstract}
Supralinear and sublinear pre-synaptic and dendritic integration is considered to be responsible for nonlinear computation power of biological neurons, emphasizing the role of nonlinear integration as opposed to nonlinear output thresholding. How, why, and to what degree the transfer function nonlinearity helps biologically inspired neural network models is not fully understood. Here, we study these questions in the context of echo state networks (ESN). ESN is a simple neural network architecture in which a fixed recurrent network is driven with an input signal, and the output is generated by a readout layer from the measurements of the network states. ESN architecture enjoys efficient training and good performance on certain signal-processing tasks, such as system identification and time series prediction. ESN performance has been analyzed with respect to the connectivity pattern in the network structure and the input bias. However, the effects of the transfer function in the network have not been studied systematically. Here, we use an approach $\tanh$ on the Taylor expansion of a frequently used transfer function, the hyperbolic tangent function,   to systematically study the effect of increasing nonlinearity of the transfer function on the memory, nonlinear capacity, and signal processing performance of  ESN. Interestingly, we find that a quadratic approximation is enough to capture the computational power of  ESN with $\tanh$ function. The results of this study apply to both software and hardware implementation of ESN.
\end{abstract}
\IEEEpeerreviewmaketitle

\section{Introduction}

McCullough and Pitts \cite{McCulloch1943} showed that the computational power of the brain can be understood and modeled at the level of a single neuron. Their simple model of the neuron consisted of linear integration of synaptic inputs followed by a threshold nonlinearity. Current understanding of neural information processing reveals that the role of a single neuron in processing input is much more complicated than a linear integration-and-threshold process  \cite{koch2000}. In fact, the morphology and physiology of the synapses and dendrites create important nonlinear effects on the spatial and temporal integration of synaptic input into a single membrane potential \cite{magee200}. Moreover,  dendritic input integration in certain neurons may adaptively switch between supralinear and sublinear regimes \cite{Margulis1998}. From a theoretical standpoint this nonlinear integration is directly responsible for the ability of neurons to classify linearly inseparable patterns \cite{10.1371/journal.pcbi.1002867}. The advantage of nonlinear processing at the level of a single neuron has also been discussed in the artificial neural network (ANN) community \cite{Giles:87}.




Historically, the ANN community has been  concerned with  algorithms for finding the correct interaction pattern between neurons for a specific task \cite{Rosenblatt:1958p1436,rumelhart1986,Hopfield:1982p652}. Some work in the field has emphasized the importance of suitable collective behavior of the neural network facilitated by macroscopic parameters over microscopic degrees of freedom. Dominey et al. \cite{Dominey:1995qo} proposed a simple model for the context-dependent motor control of eyes. In this model, the prefrontal cortex represents a suitable high-dimensional mapping of visual input that is adaptively projected onto basal ganglia, which in turn control the eye movement. The only task-dependent learning in this model occurs in the projection layer. This model has also been used to explain higher-level cognitive tasks such as grammar comprehension in the brain \cite{10.1371/journal.pone.0052946}. 

\begin{figure*}[!th]
\centering
\includegraphics[width=5in]{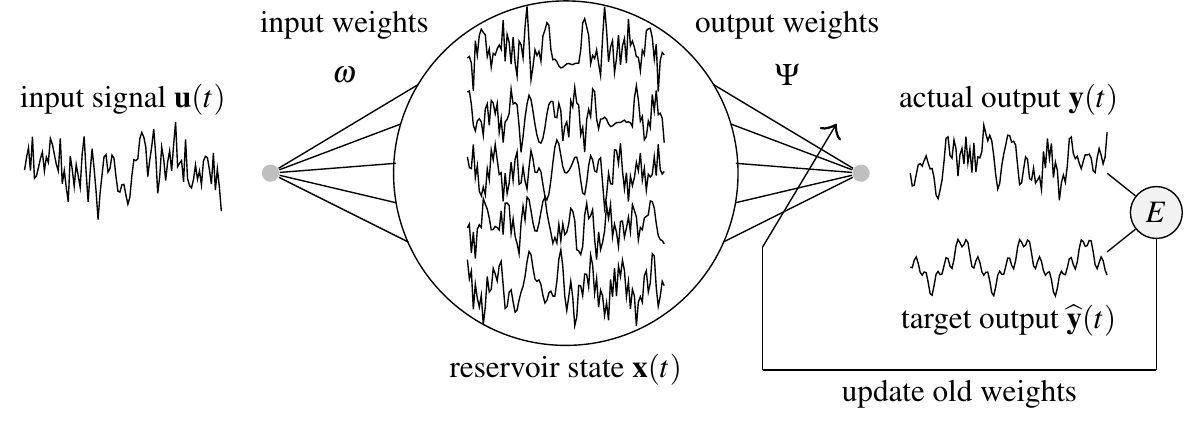}
\caption{Computation in an ESN. The reservoir is an excitable recurrent network with $N$ readable output states represented by the vector ${\bf X}(t)$. The input signal ${\bf u}(t)$ is fed into one or more points $i$ in the reservoir with a corresponding weight $\boldsymbol\omega_i$, denoted by the weight column vector ${\boldsymbol\omega}=[\boldsymbol\omega_i]$.}
\label{fig:rcdyn}
\end{figure*}

More abstract versions of this model, Liquid State Machines \cite{Maass:2002p1444} and Echo State Networks \cite{Maass:2002p1444,Jaeger02042004}, were later introduced in the neural network community  and were subsequently unified under the name {\em reservoir computing} (RC) \cite{verstraeten2007}. In RC, an easily tunable high-dimensional recurrent network, called the reservoir, is driven by an input signal. An adaptive readout layer then combines the reservoir states to produce a desired output. Figure~\ref{fig:rcdyn} provides a conceptual illustration of RC. ESN implements this idea with a discrete-time recurrent network with linear or $\tanh$ activation functions and a linear readout layer  trained using regression. Many variations of ESN exist and have been successfully applied to different engineering tasks, such as time series prediction and system identification \cite{springerlink:10.1007}. 

Owing to its fixed recurrent connections, training an ESN is much more efficient than ordinary recurrent neural networks (RNN), making it feasible to use its power in practical applications. ESN's power in time series processing has been attributed to the reservoir's memory \cite{PhysRevLett.92.148102,Dambre:2012fk} and high-dimensional projection of the input which acts like a temporal discriminant kernel \cite{Hermans:2011fk} that is present in the critical dynamical regime, where input perturbations in the reservoir dynamics neither spread nor die out \cite{Bertschinger:2004p1450,PhysRevE.87.042808,4905041020100501}.
 
A major research direction in RC is to study how the nonlinear dynamics of the reservoir may improve the performance in different tasks \cite{springerlink:10.1007,goudarzi2015a}. In particular, the goal is to understand and enhance the high-dimensional nonlinear mapping created by the reservoir dynamics. In the case of ESN architecture, the nonlinearity of the reservoir can be ascribed to its connectivity pattern, transfer function, and the input bias. While there have been some studies focusing on the effect of connectivity and bias \cite{4905041020100501,5596492},  the transfer function nonlinearity has never been systematically studied, to the best of our knowledge.

Here, we examine what happens when we replace the $\tanh$ function in the ESN reservoir with its partial Taylor series expansion, varying the number of terms included. The addition of each successive term will increase the order of nonlinearity present in the transfer function,  allowing us to gradually interpolate between the linear and the $\tanh$ transfer functions. In addition, we will explore the input weight scaling to study the effect of sublinear integration on ESN performance, at each level of nonlinearity. To control for  other sources of variation, we will restrict ourselves to the two most constrained reservoir architectures that are known to preserve the computational performance of the classical random reservoir, the simple cycle reservoir (SCR) \cite{5629375} and the Gaussian orthogonal reservoir \cite{PhysRevLett.92.148102}.

The main contribution of this work is a systematic study of the role of the transfer function nonlinearity in the total information processing capacity of recurrent neural networks. Section~\ref{sec:background} outlines the context and motivation of this work. In Section~\ref{sec:standard_esn}, we review the basic ESN formulation used in this study. In Section~\ref{sec:taylorexp}, we describe the details of our Taylor expansion approach to quantify the degree of nonlinearity and its impact on the performance of $\tanh$-neuron ESN. The experimental study on information processing properties of ESNs with Taylor expanded transfer functions is presented in Section~\ref{sec:results}. We first study the memory and also the nonlinear memory capacity of echo state networks with different transfer function nonlinearity, then we evaluate the performance of such networks against time-series  tests of Mackey-Glass and NARMA 10. In all cases, we find that the second order approximation of the $\tanh$ function provides all the nonlinear benefits of the $\tanh$ with no significant improvement to the network performance with increasing nonlinearity. Moreover, we show that the region of the $\tanh$ function which is usually thought of as linear is actually very nonlinear. RC has been suggested as a suitable signal processing framework for hardware realizations targeting unconventional substrates \cite{goudarzi2015b} and ultra-low power implementations, due to its multitasking capability, robustness to noise and variations, and a fixed computational core \cite{Goudarzi2014,Goudarzi2014a}. The result of this work can be used to simplify potential hardware designs for RC while preserving their accuracy.

\section{Background}
\label{sec:background}

\begin{figure*}[ht]
\centering
\subfloat[$\tanh$ function]{
\includegraphics[width=2.0in]{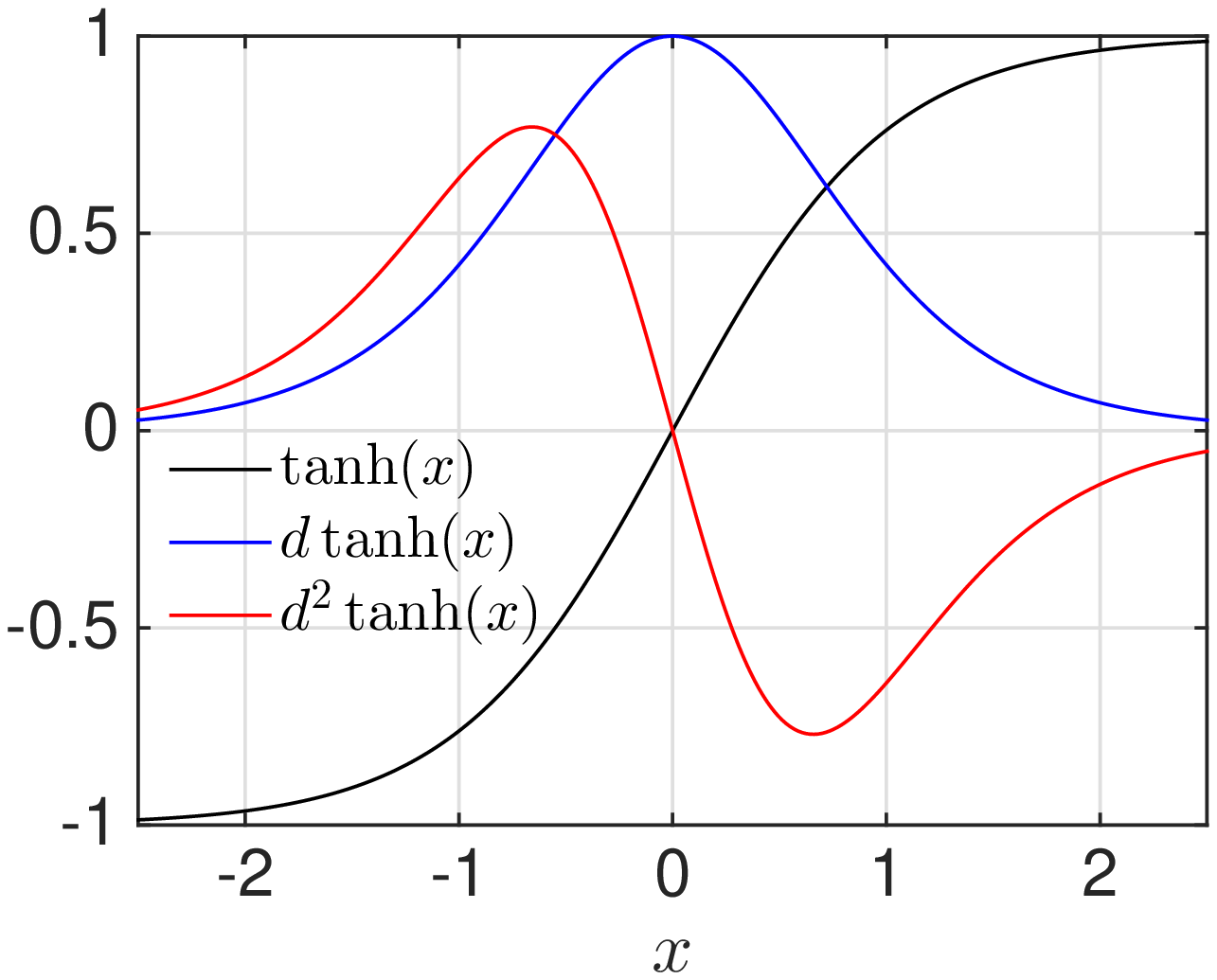}%
\label{fig:tanhfunc}
}
\subfloat[$\tanh$ Taylor series expansion]{
\includegraphics[width=2.0in]{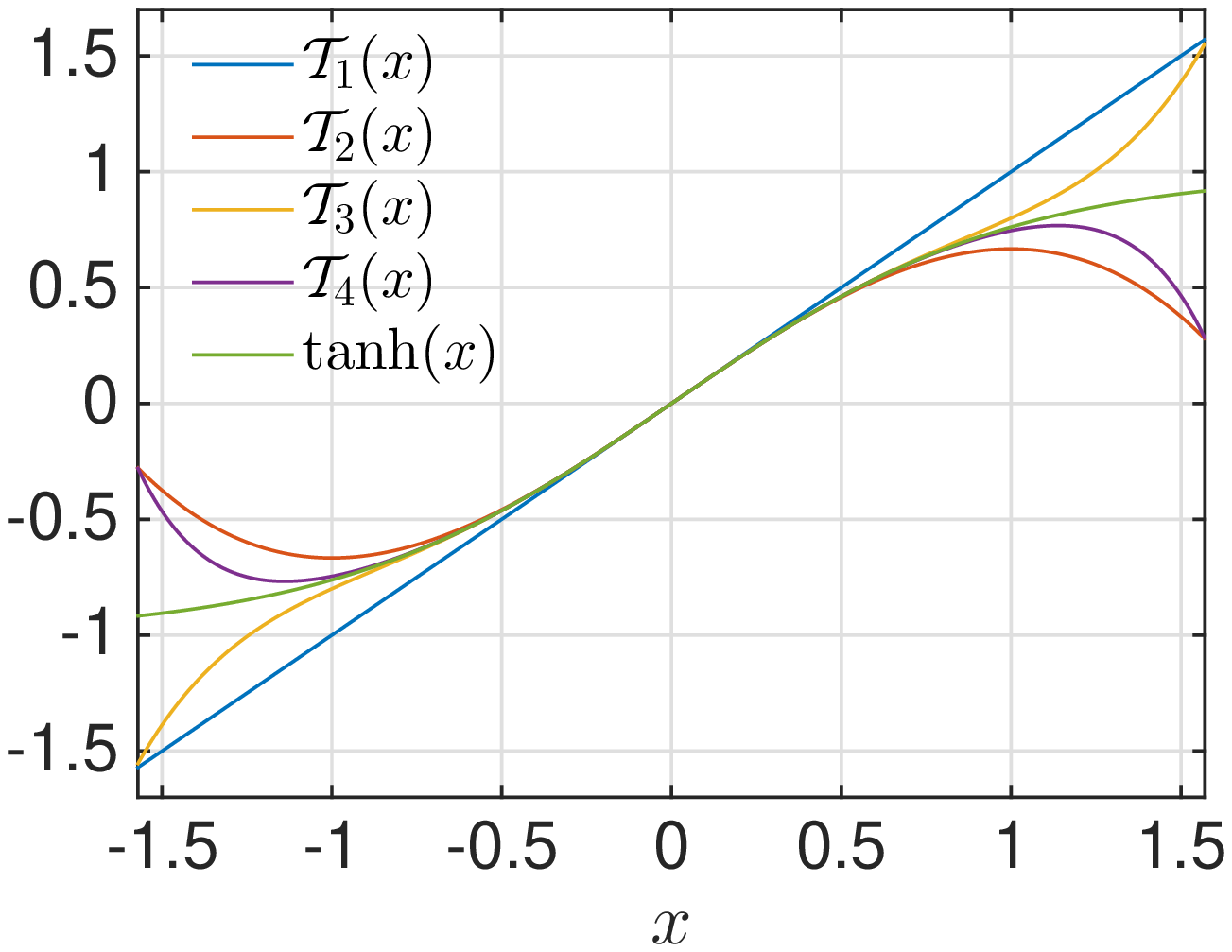}%
\label{fig:taylorseries}
}
\subfloat[approximating $\tanh$]{
\includegraphics[width=2.0in]{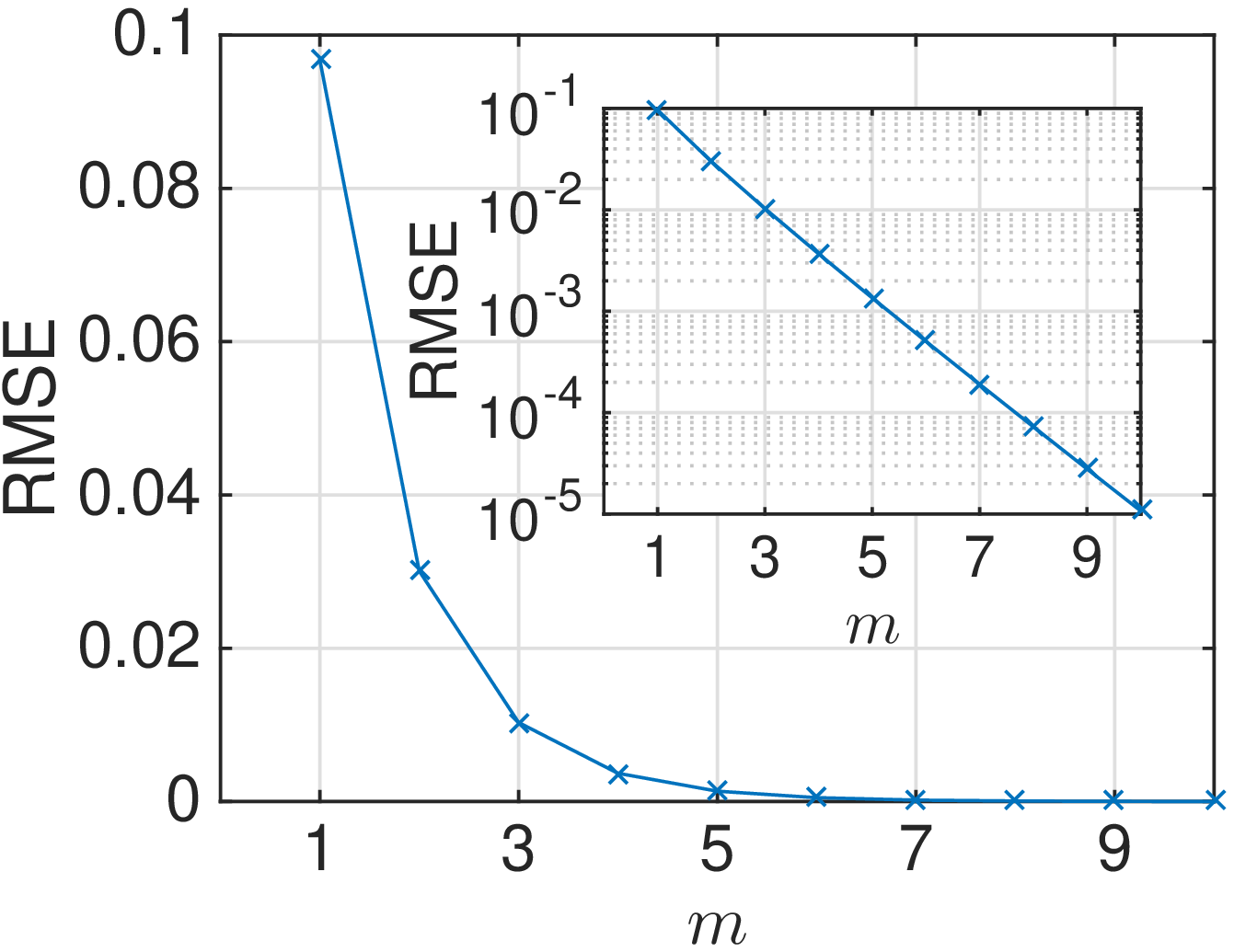}%
\label{fig:taylorerror}
}
\caption{(a) $\tanh$ and its first and second derivatives. (c) Taylor series approximation to $\tanh$ (c) Distance of Taylor series expansions to $\tanh$.
}
\label{fig:tanh}
\end{figure*}

Understanding the nature of computation and its properties is an active subject of theoretical study in reservoir computing. Hermans and Schrauwen \cite{Hermans:2011fk} showed that the ESN reservoir acts as a recursive kernel that generates a high-dimensional mapping of an input signal that can be used by the readout layer to reconstruct a target output. B{\"u}sing et al. \cite{4905041020100501} studied the relationship between the reservoir and its performance and found that while   in continuous reservoirs the performance of the system does not depend on the topology of the reservoir network, coarse-graining the reservoir state will make the dynamics and the performance of the system highly sensitive to its topology. Verstraeten et al. \cite{5596492} used a novel method to quantify the nonlinearity of the reservoir as a function of input weight magnitude. They used the ratio of the number of  frequencies in the input to  the number of frequencies  in the dynamics of the input-driven reservoir  as a proxy for the reservoir nonlinearity. 
\begin{figure}[h]
\centering
\def\w{1.6in}
\subfloat[$\tanh(x)$, memory]{
\includegraphics[width=\w]{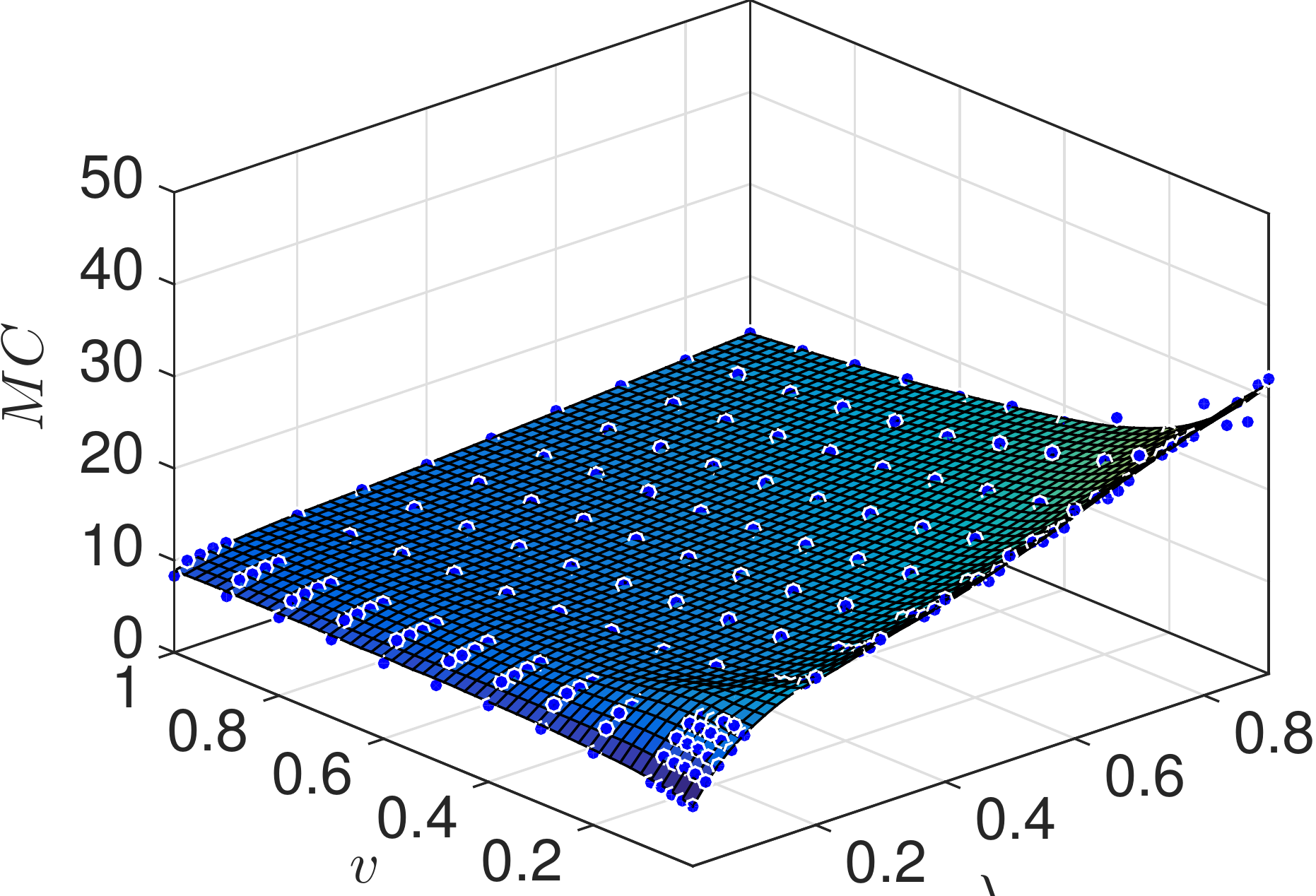}%
\label{fig:tanhmc}
}
\subfloat[$\tanh(x)$, nonlinear computation $n=3$]{
\includegraphics[width=\w]{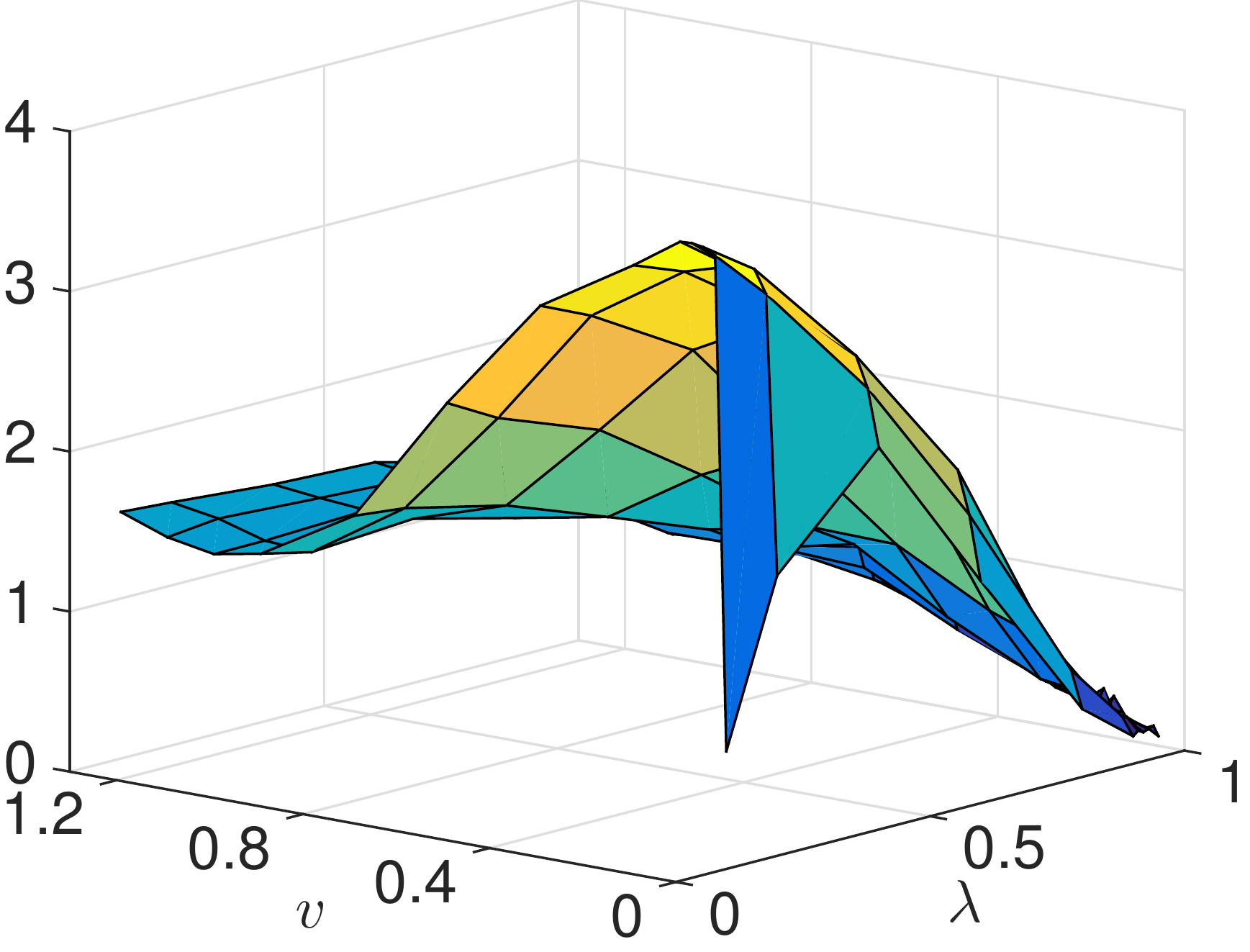}%
\label{fig:tanhnmc}
}\\
\subfloat[$\tanh(x)$, Mackey-Glass]{
\includegraphics[width=\w]{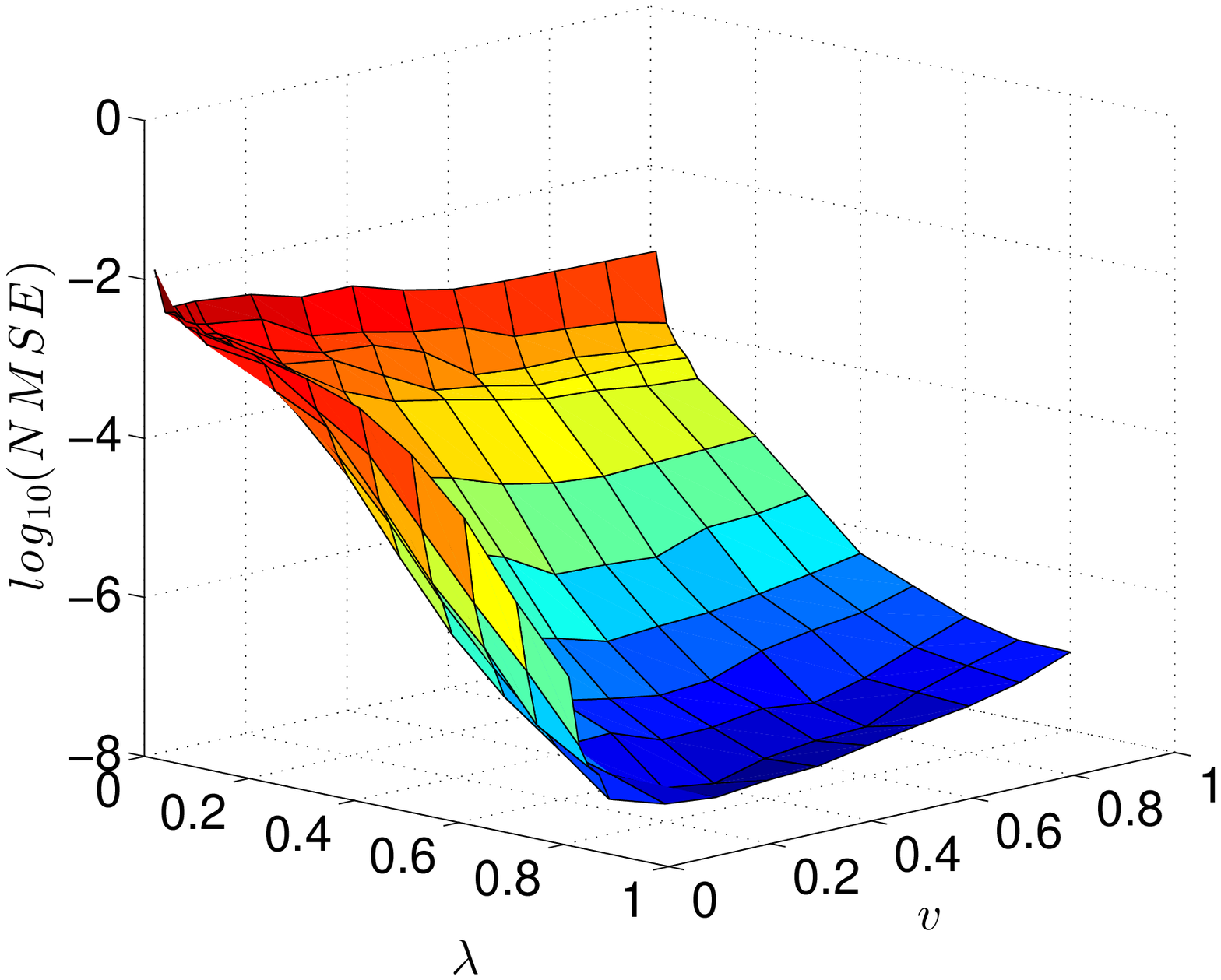}%
\label{fig:tanhmg}
}
\subfloat[$\tanh(x)$, NARMA10]{
\includegraphics[width=\w]{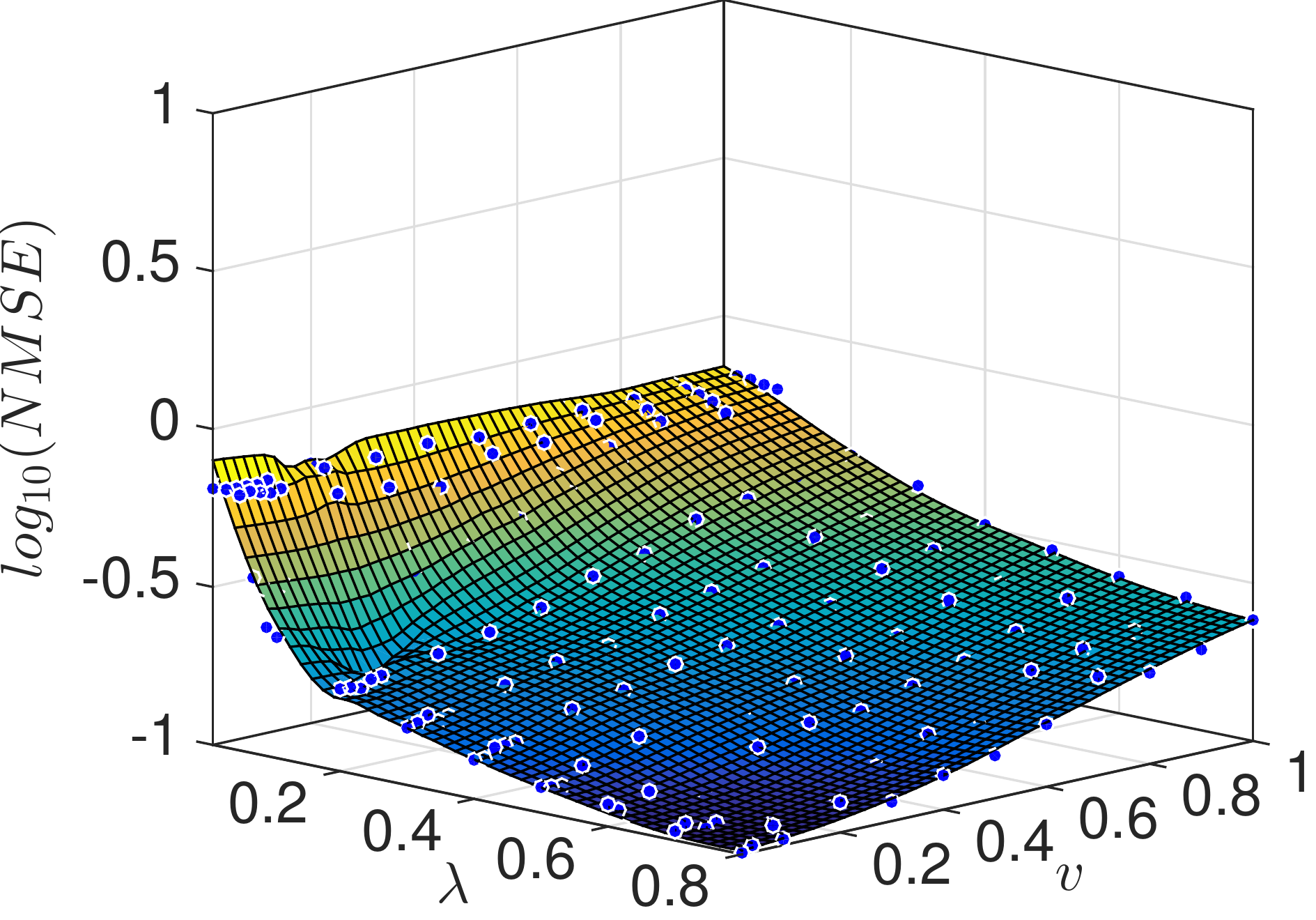}%
\label{fig:tanhnarma}
}
\caption{Illustration of sensitivity of ESN performance to $v$ and $\lambda$. For linear memory and nonlinear capacity the highest values are optimal and for Mackey-Glass prediction and NARMA 10 computation the lowest values are optimal. The optimal values for all tasks occur for low $v$, where the input signal is mapped onto the so-called linear region of the $\tanh(x)$ function.
}
\label{fig:tanhsensitivity}
\end{figure}
As a result of these studies the growing consensus is that from a theoretical perspective one would obtain more nonlinear computational power in the reservoir by adjusting the input weight magnitudes such as to project the input onto the more nonlinear regions of the $\tanh$ transfer function \cite{Butcher201376} (see Figure~\ref{fig:tanhfunc}).  This  opens an interesting research area; however, in existing approaches the linear and nonlinear regions of the $\tanh$ function are not defined precisely. Moreover, there is little evidence that using the so-called nonlinear region of the $\tanh$ actually improves the performance on nonlinear tasks\cite{Butcher201376,5629375,5596492}.

To illustrate the effect of using the nonlinear parts of the $\tanh$ function, we have included sensitivity analysis of reservoirs with linear and $\tanh$ transfer functions for solving four different benchmarks, linear memory, nonlinear computation capacity, Mackey-Glass chaotic prediction, and NARMA 10 computation (see Section~\ref{sec:results} for task details and Section~\ref{sec:standard_esn} for reservoir model). For memory and nonlinear computation capacity a reservoir of $N=50$ nodes was used, and for Mackey-Glass and NARMA 10 tasks reservoirs of $N=500$ and $N=100$ nodes were used, respectively. The reservoirs are generated by sampling the standard Gaussian distribution and are rescaled to have spectral radius $\lambda$. Input weights are drawn from the Bernoulli distribution over $\{-1,+1\}$ and multiplied by input weight coefficient $v$.  The reservoir parameters $v$ and $\lambda$ were swept on the interval $[0.1,1]$ with $0.1$ increments and the results were averaged over 10 runs.
Figure~\ref{fig:tanhsensitivity} shows the results of the sensitivity analysis. For all the tasks, the best results are achieved for the lowest $v$ values, which maps the inputs signals well within the speculated linear region of the $\tanh$ function. In this work, our goal is to decompose the nonlinearity of the $\tanh$ function and study its effects as a function of the degree of nonlinearity and input strength $v$.
\begin{figure}
\centering
\subfloat[linear reservoir]{
\includegraphics[width=2.9in]{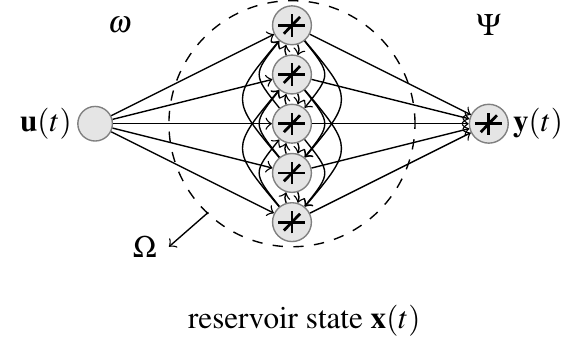}%
\label{fig:arch_esn_lin}
}\\
\subfloat[Taylor series reservoir]{
\includegraphics[width=2.9in]{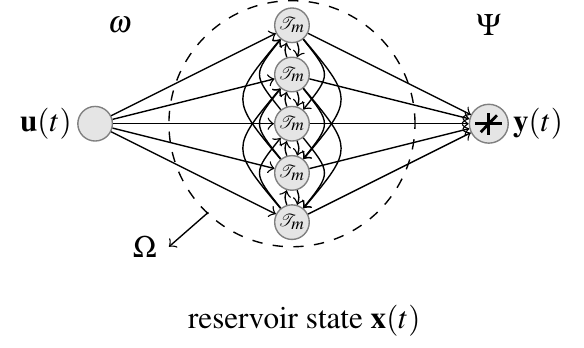}%
\label{fig:arch_esn_taylor}
}\\
\subfloat[$\tanh$  reservoir]{
\includegraphics[width=2.9in]{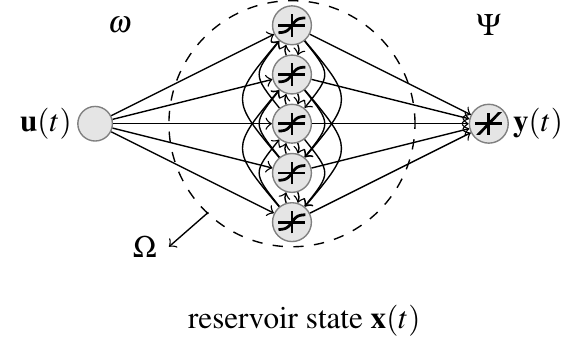}%
\label{fig:arch_esn_tanh}
}
\caption{(a) Schematic of a linear ESN. A time-varying input signal ${\bf u}(t)$ drives a dynamical core called a reservoir. The states of the reservoir ${\bf x}(t)$  are combined linearly to produce the output ${\bf y}(t)$. The reservoir consists of $N$ nodes. The input and the reservoir connections are given by the vector $\boldsymbol\omega$ and the matrix $\boldsymbol\Omega$ respectively. The reservoir states and the constant are connected to the readout layer using the weight matrix $\boldsymbol\Psi$. (b) A Taylor series ESN with a similar structure to linear ESN, but with Taylor series expansion of $\tanh$ $\tanh$ for the transfer functions of the reservoir. (c) A $\tanh$ ESN with a similar structure to linear ESN, but with $\tanh$ nonlinearity in the transfer functions of the reservoir. Usually a $\tanh$ function is used.
}
\label{fig:arch_esn}
\end{figure}

\section{Model}
\label{sec:model}
\subsection{Echo State Network}
\label{sec:standard_esn}
An ESN consists of an input-driven recurrent neural network, which acts as the reservoir, and a readout layer that reads the reservoir states and produces the output. Mathematically, the input driven reservoir is defined as follows. Let $N$ be the size of the reservoir. We represent the time-dependent inputs as a column vector $u(t)$, the reservoir state as a column vector $x(t)$, and the output as a column vector $y(t)$. The input connectivity is represented by the matrix $\boldsymbol\omega$ and the reservoir connectivity is represented by an $N\times N$ weight matrix $\boldsymbol\Omega$. For simplicity, we assume that we have one input signal and one output, but the notation can be extended to multiple inputs and outputs. The time evolution of the reservoir is given by:
\begin{equation}
x(t+1) = f(\boldsymbol\Omega x(t) + \boldsymbol\omega u(t)).
\end{equation}
where $f$ is the transfer function of the reservoir nodes that is applied element-wise to its operand. This function is usually $\tanh$ or linear. The output is generated by the multiplication of  a readout weight matrix $\boldsymbol\Psi$ of length  $N+1$ and the reservoir state vector $x(t)$ extended by an optional constant $1$ represented by $x'(t)$:
\begin{equation}
y(t) = \boldsymbol\Psi   x'(t).
\end{equation}

The readout weights $\boldsymbol\Psi$ need to be trained using a teacher input-output pair. A popular training technique is to use the pseudo-inverse method\cite{verstraeten2007}. One  drives the ESN with a teacher input and records the history of the reservoir states into a matrix ${\bf X}$, where the columns correspond to the reservoir nodes and each row  gives the states of all reservoir nodes at one time. A constant column of 1s is added to ${\bf X}$ to serve as a bias. The corresponding teacher output will be denoted by the column vector ${\bf \widehat{y}}$. The readout can be calculated as follows:
\begin{equation}
\Psi = \langle {\bf XX'}\rangle^{-1} \langle {\bf X} {\bf \widehat{Y}'}\rangle,
\label{eq:regression}
\end{equation}
where $'$ indicates the transpose of a matrix. Figures~\ref{fig:arch_esn_lin} and~\ref{fig:arch_esn_tanh} show the architecture of ESNs with linear and $\tanh$ activation functions, respectively. Figure~\ref{fig:arch_esn_taylor} shows the architecture of an ESN with the Taylor series approximation of $\tanh$ as transfer function. In the next section, we will describe how we will use these approximations to systematically study the transfer function nonlinearity in the reservoir.

\subsection{Transfer Function Nonlinearity}
\label{sec:taylorexp}
Our goal is to systematically explore the  effect of nonlinearity of the reservoir transfer function on the ESN memory and performance. Figure~\ref{fig:tanhfunc} illustrates the $\tanh(x)$ function and its first and second derivatives, i.e., $d \tanh(x)$ and $d^2 \tanh(x)$. The $\tanh(x)$ function is often considered to behave linearly for $|x|<0.5$, and nonlinearly otherwise. However, looking closely at the curves of $d\tanh(x)$ and $d^2 \tanh(x)$, we see that the only place where the $\tanh(x)$ behaves linearly (constant $d\tanh(x)$) is when
$x\to 0$. As $x$ increases in magnitude its first derivative changes very rapidly with increasing rate, i.e., steep $d^2 \tanh(x)$, until $|x|=0.65$. This observation suggests that the so-called linear region of $\tanh(x)$ function is where the function becomes highly nonlinear very quickly as $x$ increases.

We would like to decompose the nonlinearity of $\tanh$ and study how much each additional degree of  nonlinearity affects the performance of the ESN. To this end, we use the Taylor series expansion of the  $\tanh$  function around $x=0$ to systematically interpolate the orders of nonlinearity between the linear transfer function to $\tanh$ transfer function. We will replace the $\tanh$ transfer function with the transfer functions that we obtain by writing the $\tanh$ Taylor series to $m$ terms, denoted by $\mathcal{T}_m$. Table~\ref{tab:tanhexpansion} lists the first few expansions as well as the exact Taylor series for $\tanh$ and Figure~\ref{fig:arch_esn_taylor} illustrates the architecture of the ESN with Taylor series expansions as the reservoir transfer function. 
\begin{table}[h]
\centering
\begin{tabular}{c|c}
\hline
 $m$ & expansion \\ \hline
 $\mathcal{T}_1(x)$ & $x$\\
 $\mathcal{T}_2(x)$ & $x - \frac{1}{3} x^3$\\
 $\mathcal{T}_3(x)$ & $x - \frac{1}{3} x^3 + \frac{2}{15} x^5$\\
 $\mathcal{T}_4(x)$ & $x - \frac{1}{3} x^3 + \frac{2}{15} x^5 - \frac{17}{315}x^7$\\
  \vdots & \vdots\\
  $\tanh(x)=\mathcal{T}_\infty(x)$ & $\sum_{m=1}^\infty \frac{B_{2m}4^m(4^m-1)}{(2m)!}x^{2m-1}$\\ \hline
\end{tabular}
\caption{Example of Taylor series expansions of $\tanh(x)$ with different orders $m$. Here, $B_{2m}$ is the number at the position $2m$ in Bernoulli sequence.}
\label{tab:tanhexpansion}
\end{table}

Figure~\ref{fig:taylorseries} shows the curves corresponding to the first four expansions of the $\tanh(x)$ function. Although the Taylor series expansion of $\tanh(x)$ is defined for $|x|<\pi$, it is only for $|x|<1$ that the lowest order expansions do not rapidly diverge from $\tanh(x)$. Figure~\ref{fig:taylorerror} shows the root-mean-squared error (RMSE) between the Taylor expansion $m$ and the $\tanh(x)$ function calculated for $|x|<1$. With increasing number of terms in the expansion, the approximation approaches the true $\tanh$ exponentially (the inset plot). Understanding this exponential behavior suggests most of the benefits of the $\tanh$ nonlinearity may come from the first few orders of nonlinearity, and this  will help us to interpret the results in the later sections.

\section{Experiments}
\label{sec:results}

In this section we study the effect of nonlinearity of the transfer function in ESNs using two parameters, the input weight coefficient $v$ and the order of the Taylor series expansions used as the transfer function $m$. We will evaluate the performance of ESNs in linear memory capacity, nonlinear capacity, Mackey-Glass chaotic time series prediction, and NARMA 10 computation.

To make a fair comparison between systems, we adjust $v$ and the input signal scaling so that the the magnitude of the reservoir states is less than 1. The next section will give the details of ESN construction and evaluation.

\subsection{Reservoir Construction and Evaluation}

To control for the variations that are due to topological factors, we will use very constrained reservoir architectures. For the memory task, Mackey-Glass prediction, and NARMA 10 computation we will use the simple cycle reservoir \cite{5629375}. This topology compares well with random topology in memory and signal-processing benchmark performance, while minimizing the structural variations of the reservoir. In the simple cycle reservoir, the reservoir is a simple ring topology with uniform positive weights $r$. In this topology the weight $r$ determines the reservoir spectral radius: $r=|\lambda|$ and no rescaling of the weight matrix is needed. In initial experiments, we observed that the simple cycle is unable to perform the nonlinear capacity task. For this task we create the reservoir by sampling the Gaussian orthogonal ensemble (GOE) \cite{PhysRevLett.92.148102}. The reservoir weight matrix in this case is given by ${\boldsymbol\Omega} = A + A'$, where $A$ is a matrix with the same dimensionality as ${\boldsymbol\Omega}$ where the entries are sampled from the standard Gaussian distribution $\mathcal{N}(0,1)$. The reservoir is then rescaled to have spectral radius $\lambda$. The number of reservoir nodes $N$ is adjusted for each task to get reasonably good results in a reasonable amount of time. The input weights are generated by sampling the Bernoulli distribution over $\{-1,+1\}$ and multiplying with the input weight coefficient $v$.  The reservoir nodes are initialized with $0$s and a washout period of $2N$ is used during training and testing.

The reservoirs are driven with task-dependent input $u_t$ for $2,000$ time steps and the readout weights $\Psi$ are calculated as described in Section~\ref{sec:standard_esn} using MATLAB's {\em pinv()} function. For evaluation, the reservoir state is reinitialized and the reservoir is driven for another $T=2,000$ time steps and the output $y_t$ is generated. For brevity, throughout the experiments section we adopt the subscript notation for the time index, e.g., $y_t$ instead of $y(t)$. By convention, the system performance for computational capacity tasks is evaluated using the capacity function $C_\tau$, which is the coefficient of determination between the output $y_t$ and the desired output $\widehat{y}_t$:
\begin{equation}
C_{\tau} = \frac{\mathrm{Cov}^2(y_t,\widehat{y}_t)}{\mathrm{Var}(y_t)\mathrm{Var}(\widehat{y}_t)},
\end{equation}
where $\tau$ is the memory length for the task (see Section~\ref{sec:membench} for details). For the chaotic prediction task, the performance is evaluated by calculating the normalized mean-squared-error $NMSE$ as follows:
\begin{equation}
NMSE = \frac{\sqrt{ \frac{1}{T} \sum_{t=0}^{T} (y_t - \widehat{y}_t)^2}}{\mathrm{Var}(\widehat{y}_t)},
\label{eq:nmse}
\end{equation}
where $y_t$ is the network output and $\widehat{y_t}$ is the desired output.

For all tasks we systematically explore $v\in\{10^{-5},\dots,10^{-1}\}$ with quarter decade increments and $v\in\{0.2,\dots,0.35\}$ with $0.05$ increments. All results are averaged over 10 runs. We chose this range for $v$ in preliminary runs in combination with appropriate input scaling for each task to ensure that the magnitude of reservoir states is always less than 1.

\subsection{Linear Memory Capacity}
\label{sec:membench}
The linear memory capacity is a standard measure of memory in recurrent neural networks. The $\tau$-delay memory function $C_\tau$  measures how long a network can remember its inputs.  These capacities are calculated by summing the capacity function over $\tau$: $C=\sum_\tau C_\tau$. We use $1\le\tau\le100$ for our empirical estimations. In these sets of experiments reservoirs of size $N=50$ nodes are driven with a one-dimensional input drawn from  uniform distributions on  $[-0.5,0.5]$. We fix $\lambda=0.9$ for all experiments. The desired output for this task is defined as:
\begin{equation}
\widehat{y}_t = u_{t-\tau}.
\end{equation}

\begin{figure}[h]
\centering
\subfloat[]{
\includegraphics[width=3in]{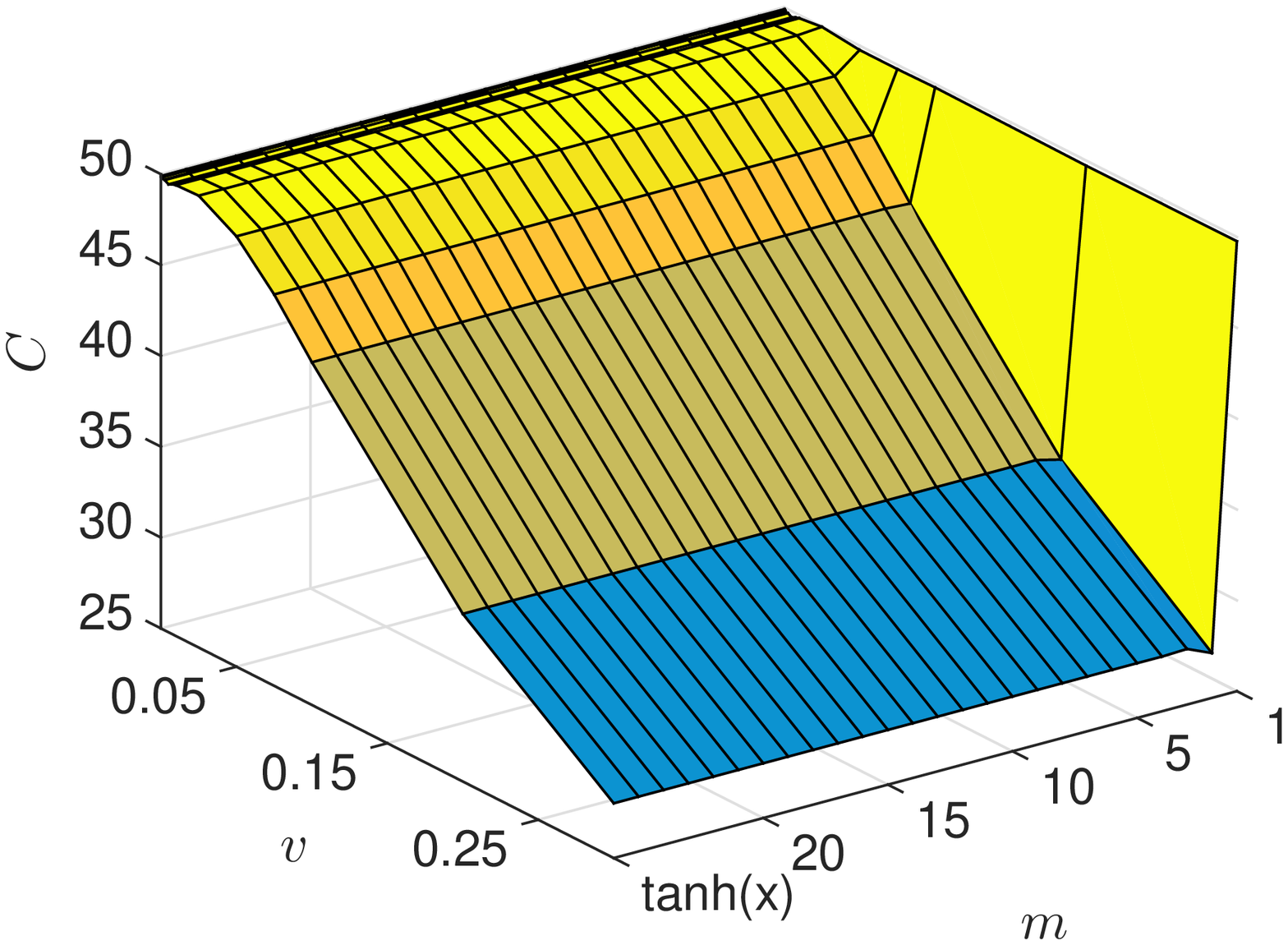}
\label{fig:mcsurf}}\\
\subfloat[]{
\includegraphics[width=1.7in]{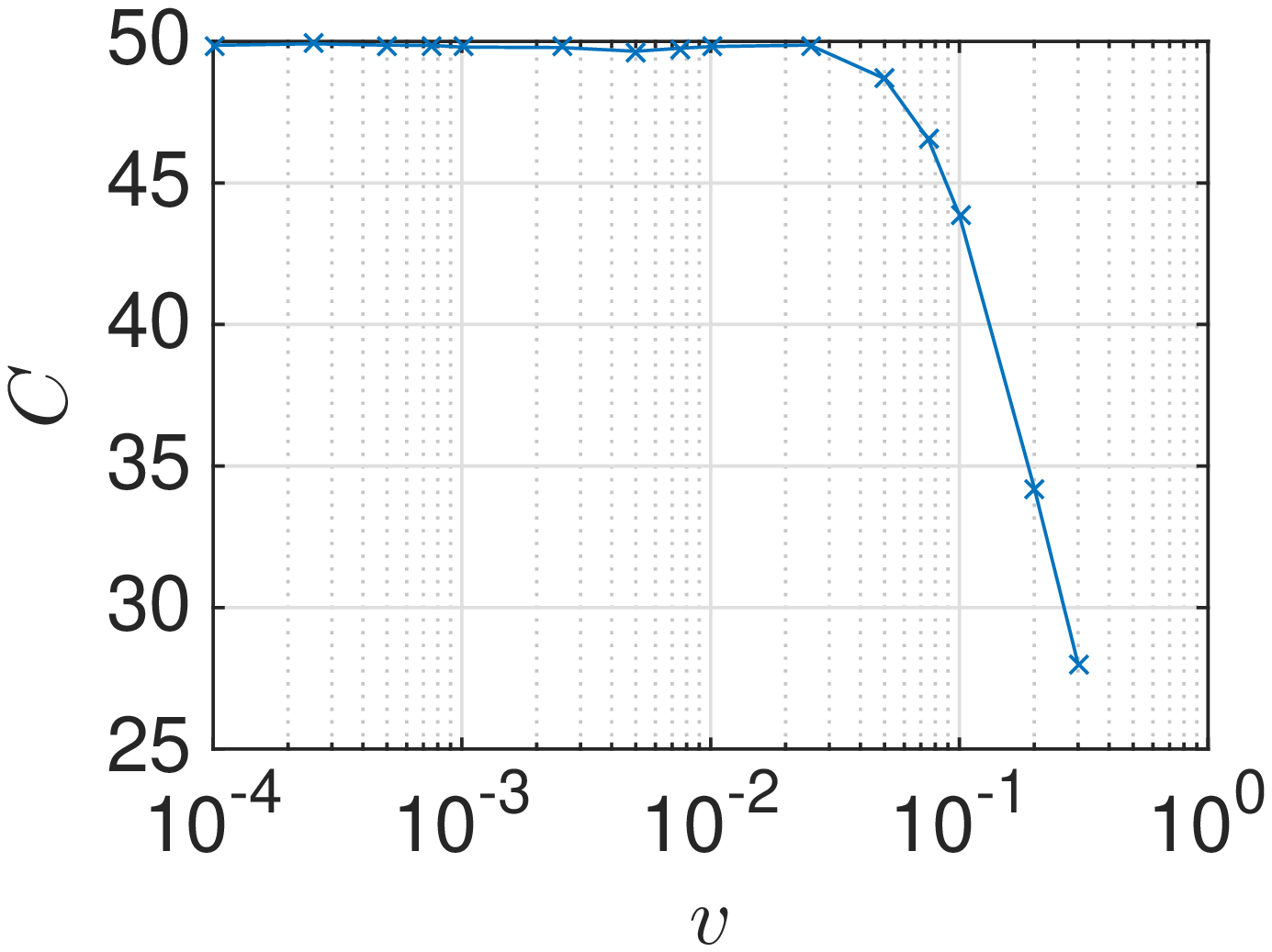}
\label{fig:mctanhv}}
\subfloat[]{
\includegraphics[width=1.7in]{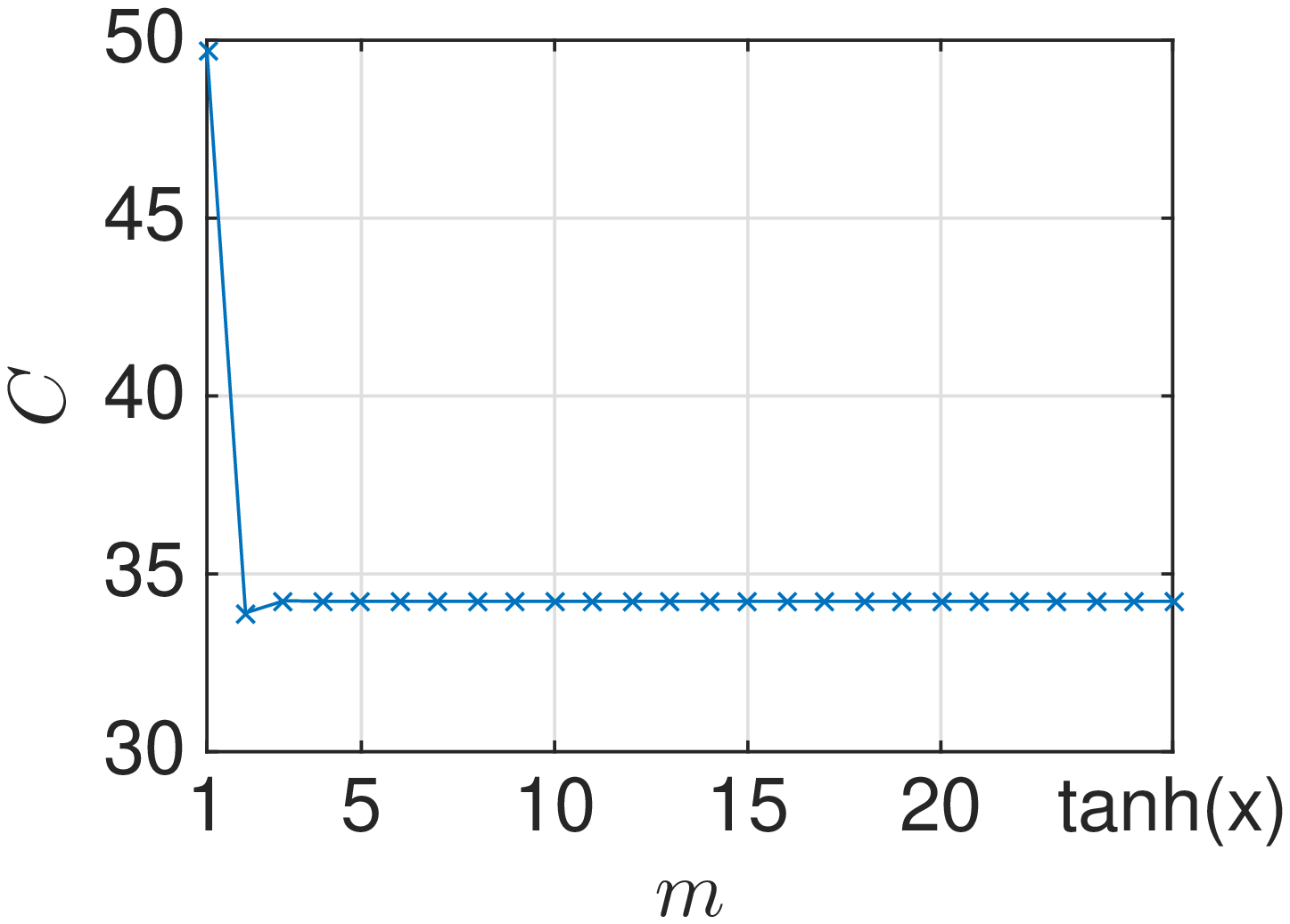}
\label{fig:mctaylorv01}}
\caption{(a) Linear memory capacity for different $v$ and $m$. (b) For $v<0.05$ the memory capacity of the $\tanh$ network is similar to that of a linear network. (c) Increasing nonlinearity beyond $m>2$ there is no change in the memory capacity of the network.}
\label{fig:linmem}
\end{figure}

Figure~\ref{fig:mcsurf} shows the total linear memory capacity surface as a function of $m$ and $v$. Consistent with previous theoretical and experimental results the linear memory capacity does not show any dependency on $v$ for $m=1$, i.e., for the linear network. However, for large $v>0.05$ and $m>1$ we observe a deviation from  linear memory with no dependence on $m$. Figure~\ref{fig:mctanhv} shows the total memory capacity for the $\tanh$ transfer function as a function of $v$ on a linear-log scale, clearly showing that for $v<0.05$ the total memory capacity of the network equals that of a linear network. Figure~\ref{fig:mctaylorv01} shows the total capacity for $v=0.1$ for various $m$, confirming that for $m>1$ the memory capacity does not vary with $m$, and suggesting that all the relevant nonlinear characteristics of the network stemming from $\tanh$ can be observed on the second-order Taylor expansion $m=2$.

\subsection{Nonlinear Computation Capacity}
The nonlinear computation capacity measures the ability of the system to reconstruct a nonlinear function of its past inputs. Commonly, Legendre polynomials are used to calculate the nonlinear computation capacity of the reservoir \cite{Dambre:2012fk}; their advantage is that Legendre polynomials of different orders are orthogonal to each other, allowing one to measure the reservoir's capacity to compute functions of varying degrees of nonlinearity independently from each other. These capacities are calculated by summing the capacity function over $\tau$: $C=\sum_\tau C_\tau$. We use $1\le\tau\le100$ for our empirical estimations. In these sets of experiments reservoirs of size $N=50$ nodes are driven with a one-dimensional input drawn from  uniform distributions on  $[-1,1]$. We fix $\lambda=0.1$ for all experiments. We have previously observed this is the optimal $\lambda$ for this task. The desired output of the Legendre polynomial of order $n$ with delay $\tau$  is given by: 
\begin{equation}
\widehat{y}(n,\tau)_t = \frac{1}{2^n} \sum_{k=0}^n {n \choose k}^2 (u_{t-\tau}-1)^{n-k}(u_{t-\tau}+1)^{k}.
\end{equation}
We must point out that unlike \cite{Dambre:2012fk}, here the network has to reconstruct the output of a single polynomial and not the product of several polynomials. In this work we only focus on the case  $n=3$. For $n=1$, the nonlinear capacity measure reduces to linear memory and the $\tanh$ are unable to compute the even orders because of the input-output symmetry.

\begin{figure}[h]
\centering
\subfloat[]{
\includegraphics[width=3in]{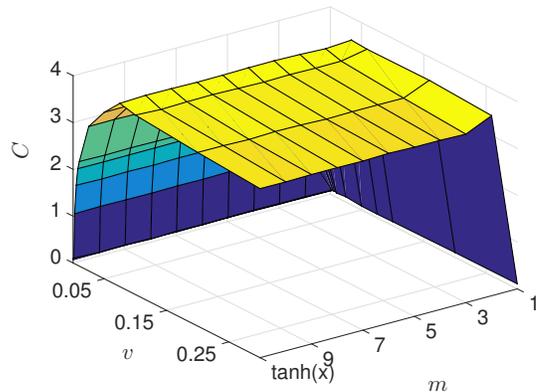}
\label{fig:nmcsurf}}\\
\subfloat[]{
\includegraphics[width=1.7in]{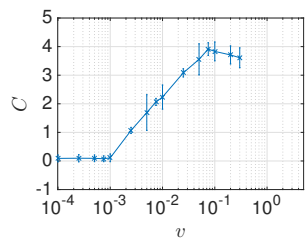}
\label{fig:nmctanhv}}
\subfloat[]{
\includegraphics[width=1.7in]{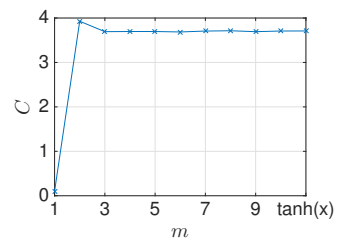}
\label{fig:nmctaylorv01}}
\caption{(a) Nonlinear capacity for different $v$ and $m$. (b) For $v<0.001$ the nonlinear capacity of $\tanh$ network is similar to that of a linear network. (c) Increasing nonlinearity beyond $m>2$ there is no change on the nonlinear capacity of the network.}
\label{fig:nmem}
\end{figure}

Figure~\ref{fig:nmcsurf} shows the total nonlinear  capacity surface as a function of $m$ and $v$. For $v>0.001$ and $m>1$ we observe a deviation from the linear network capacity, with no dependence on $m$. Figure~\ref{fig:nmctanhv} shows the nonlinear capacity for  the $\tanh$ transfer function as a function of $v$ on a linear-log scale, clearly showing that for $v<0.001$ the nonlinear capacity of the network equals that of a linear network. Figure~\ref{fig:nmctaylorv01} shows the total capacity for $v=0.1$ for various $m$, confirming that for $m>1$ the nonlinear capacity does not vary with $m$, suggesting all the relevant nonlinear characteristics of the network stemming from $\tanh$ can be observed on the second-order Taylor expansion $m=2$. We emphasize that we have used a standard ESN implementation without reservoir bias for simplicity. Applying a bias to the reservoir drastically changes the nonlinear capacity and requires a more thorough analysis.

\subsection{Mackey-Glass System Prediction}

The Mackey-Glass system \cite{Mackey15071977} is a delayed differential equation with chaotic dynamics, commonly used as a benchmark for chaotic signal prediction. This system is described by:
\begin{align}
\frac{dx_t}{dt} =  \beta \frac{x_{t-\delta}}{1+x_{t-\delta}^n} - \gamma x_t,
\end{align}
where $\beta=0.2,n=10$, and $\gamma=0.1$ are positive constants and $\delta=17$ is the feedback delay. The reservoir consists of $N=500$ nodes and $\lambda=0.9$. The task is to predict the next $\tau$ integration time steps given $x_t$. We scaled the time series between $[0,0.5]$ before feeding the network.

\begin{figure}[h]
\centering
\subfloat[]{
\includegraphics[width=3in]{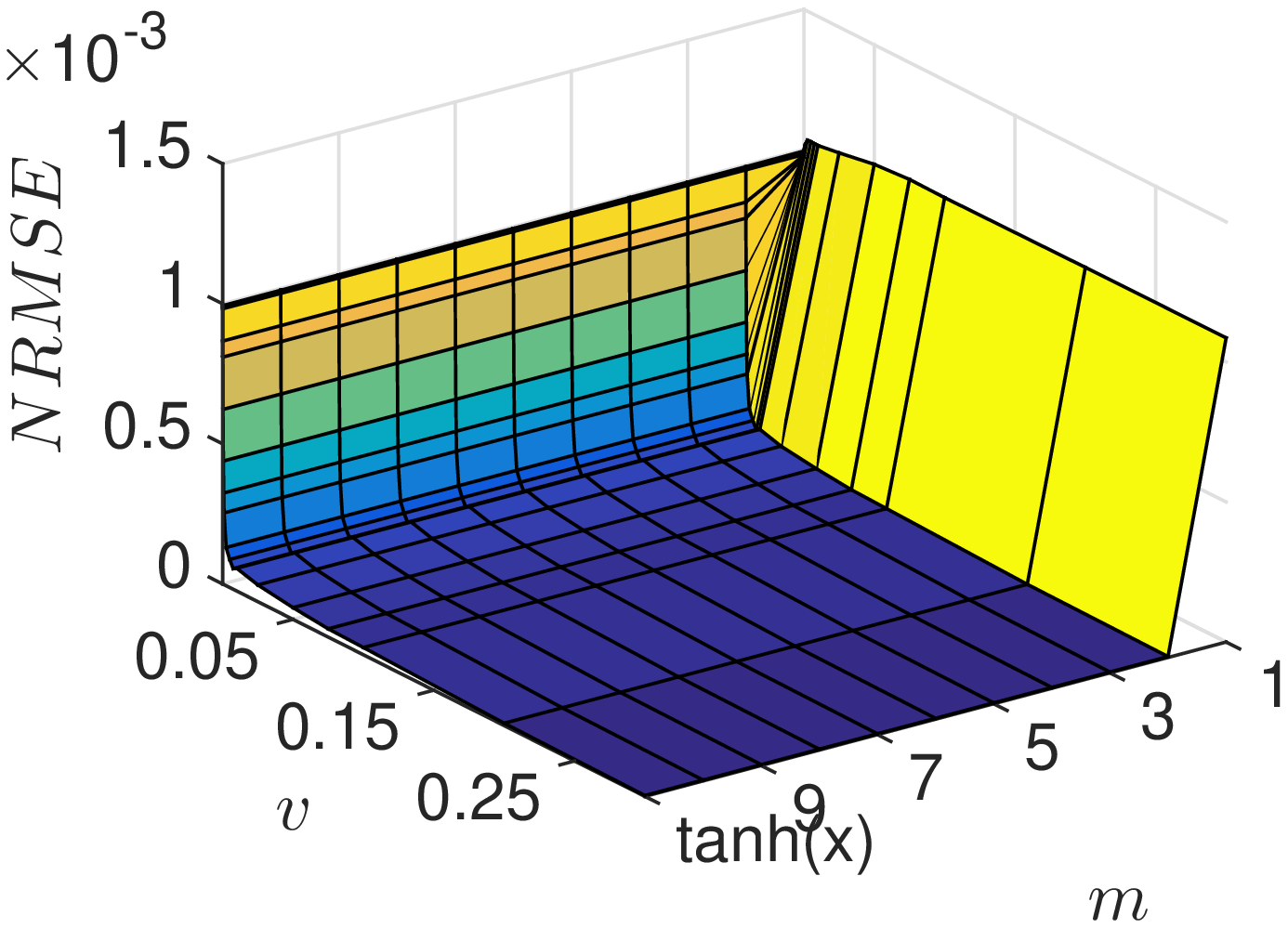}
\label{fig:mgsurf}}\\
\subfloat[]{
\includegraphics[width=1.7in]{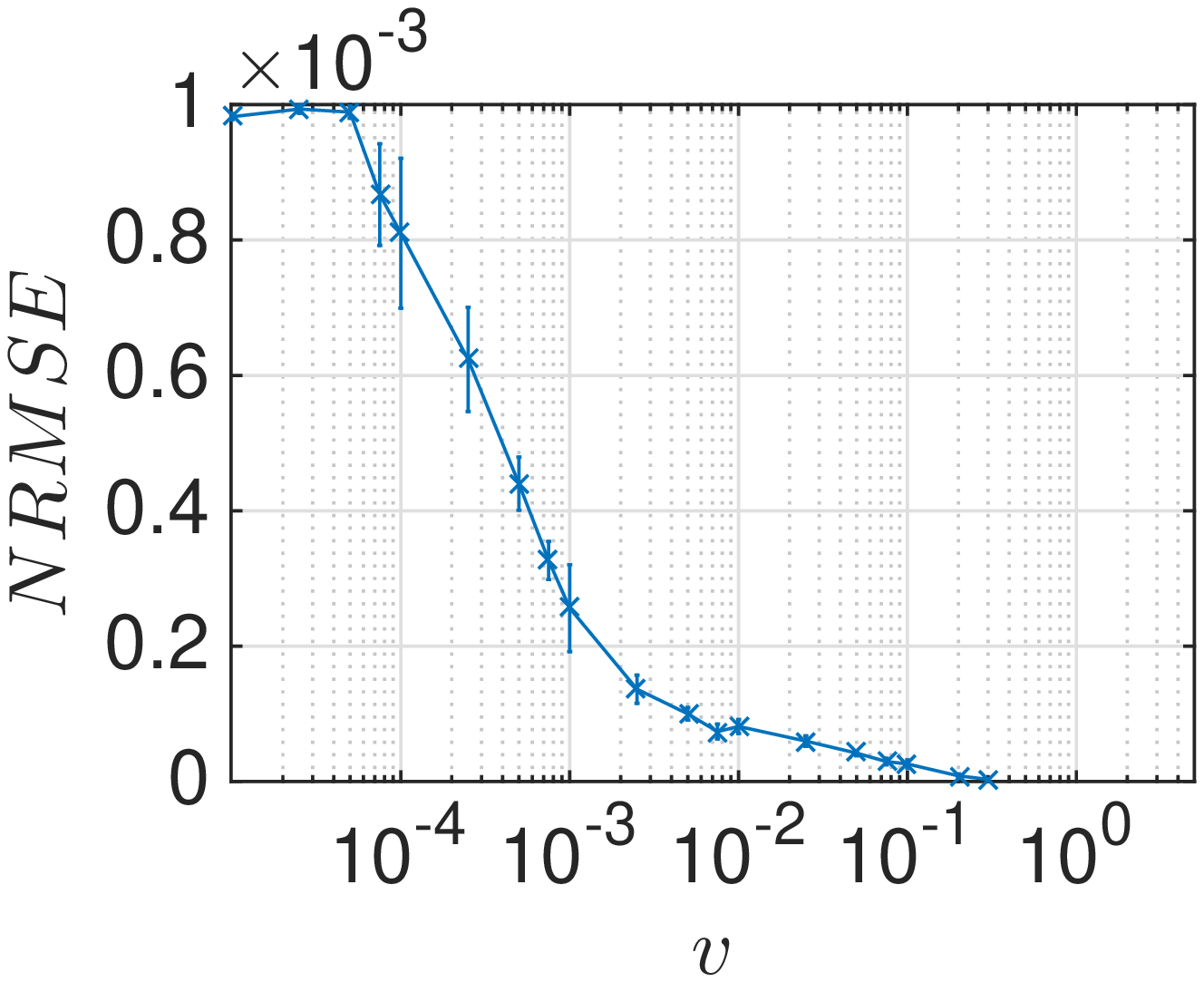}
\label{fig:mgtanhv}}
\subfloat[]{
\includegraphics[width=1.7in]{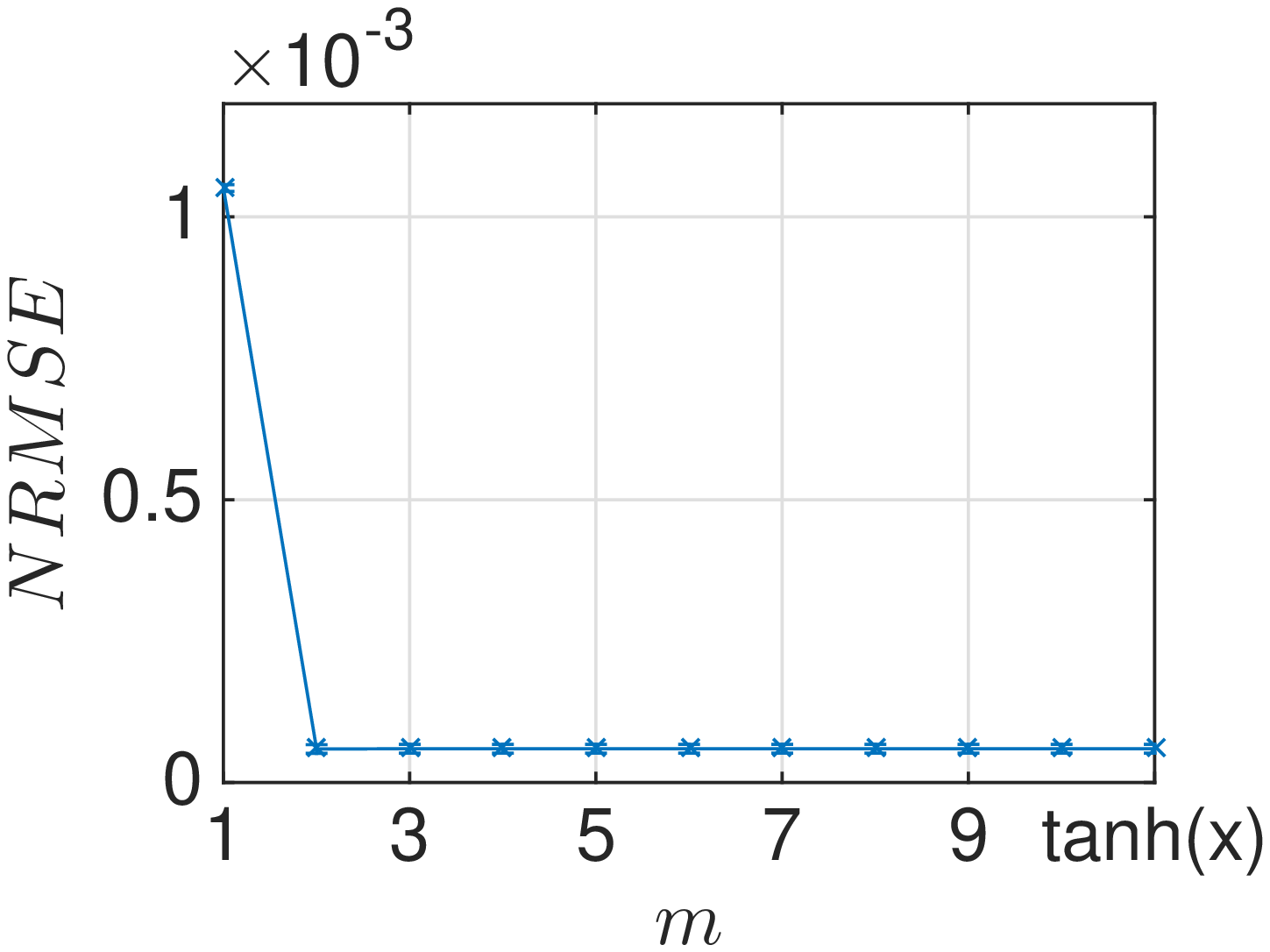}
\label{fig:mgtaylorv01}}
\caption{(a) Mackey-Glass prediction performance for different $v$ and $m$. (b) The prediction performance for $\tanh$ network. For $v<0.00075$ the the performance of $\tanh$ network is similar to that of a linear network. (c) Increasing nonlinearity beyond $m>2$ there is no change on the memory capacity of the network.}
\label{fig:MG}
\end{figure}

Figure~\ref{fig:mgsurf} shows the $NRMSE$ surface as a function of $m$ and $v$. For $m>1$ we observe a deviation from the linear network performance with no dependence on $m$. Figure~\ref{fig:mgtanhv} shows the performance for the $\tanh$ transfer function as a function of $v$ on a linear-log scale, clearly showing that for $v<0.00075$ the performance of the network equals that of a linear network, with no improvement for $v>0.1$. Figure~\ref{fig:mgtaylorv01} shows the performance for $v=0.1$ for various $m$, confirming that for $m>1$ the performance does not vary with $m$, suggesting all the relevant nonlinear characteristics of the network stemming from $\tanh$ can be observed on the second-order Taylor expansion $m=2$. In our experiments, we found that although applying a bias to the reservoir improves its nonlinear capacity, it does not improve the performance for Mackey-Glass tasks.

\subsection{NARMA 10 Computation}

 NARMA 10 \cite{5629375} is a highly non-linear auto-regressive task with long lags that is frequently used to assess neural network performance. This task is given by the following equation:  
\begin{equation}
y_t=\alpha y_{t-1}+\beta y_{t-1} \sum_{i=1}^{n}y_{t-i}+\gamma u_{t-n}u_{t-1}+\delta,
\label{eq:narma}
\end{equation}
where $n=10$, $\alpha=0.3, \beta=0.05, \gamma=1.5, \delta=0.1$. The input $u_t$ is 
drawn from a uniform distribution in the interval $[0,0.5]$. We use reservoir networks of size $N=100$ and $\lambda=0.8$. 

\begin{figure}[h]
\centering
\subfloat[]{
\includegraphics[width=3in]{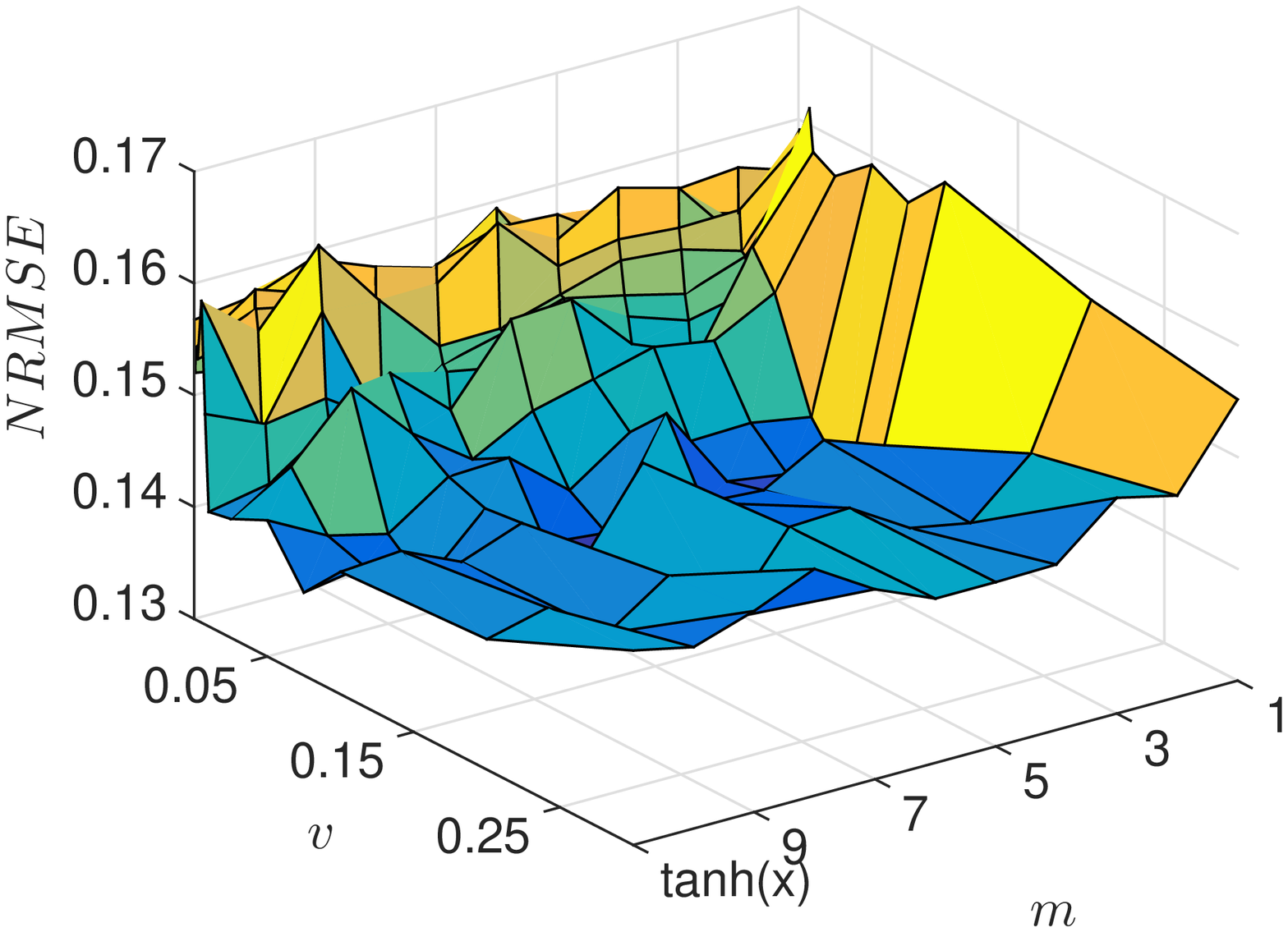}
\label{fig:narmasurf}}\\
\subfloat[]{
\includegraphics[width=1.7in]{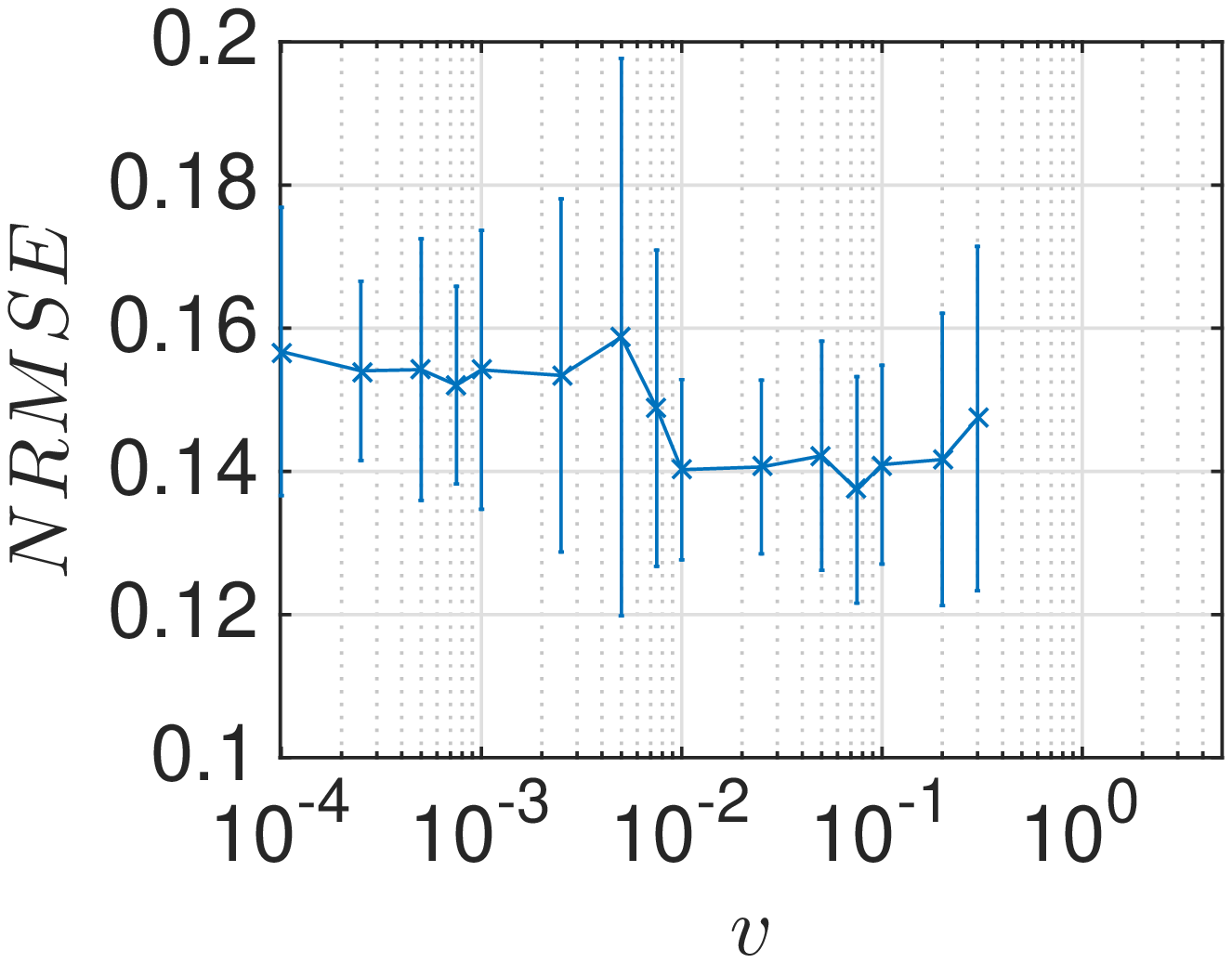}
\label{fig:narmatanhv}}
\subfloat[]{
\includegraphics[width=1.7in]{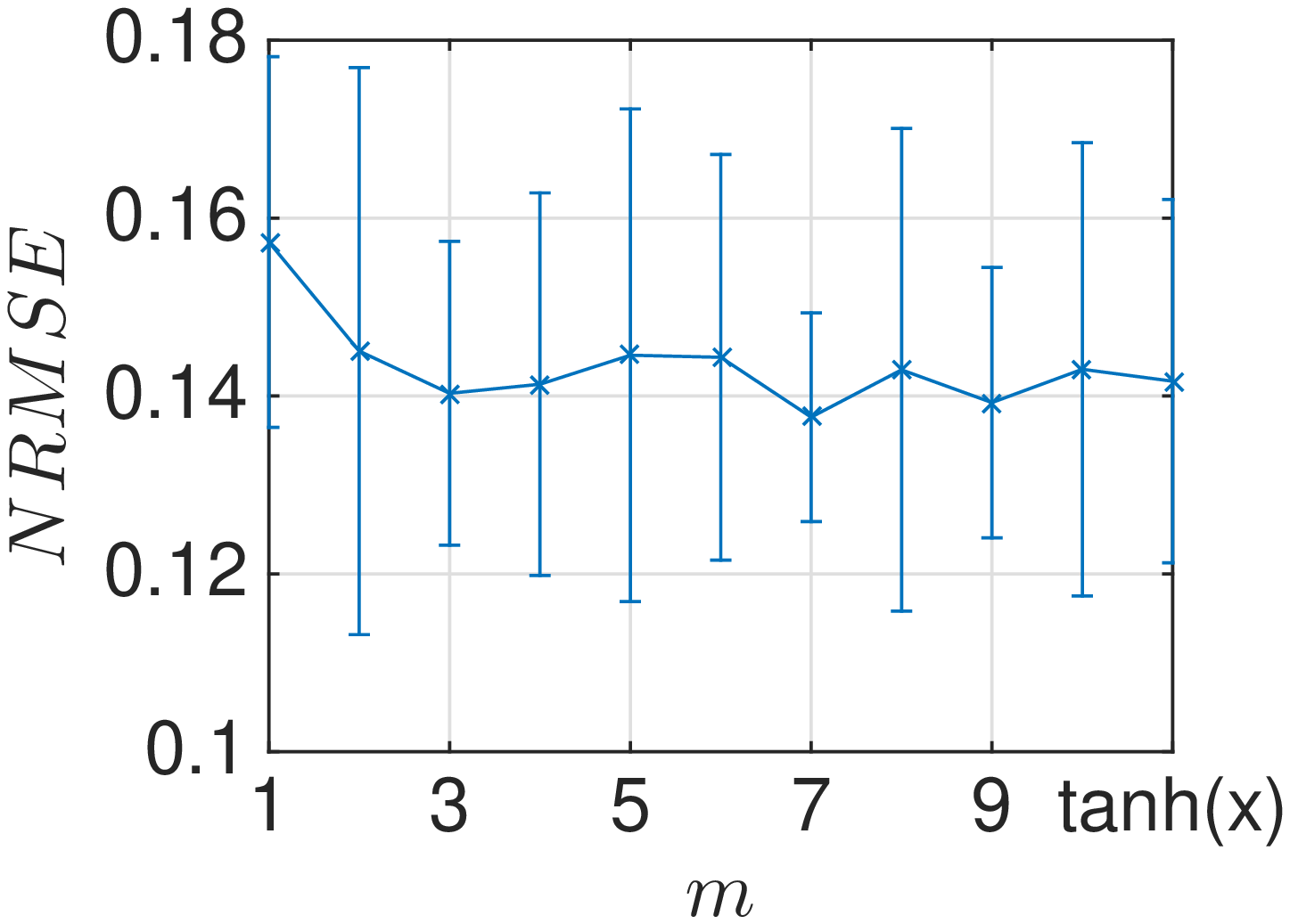}
\label{fig:narmataylorv01}}
\caption{(a) NARMA 10 performance for different $v$ and $m$. (b) The performance for $\tanh$ network. For $v<0.01$ the the performance of the $\tanh$ network is similar to that of a linear network. (c) Increasing nonlinearity beyond $m>2$ there is no change on the memory capacity of the network.}
\label{fig:narma10}
\end{figure}

Figure~\ref{fig:narmasurf} shows the $NRMSE$ surface as a function of $m$ and $v$. For $m>1$ we observe a deviation from the linear network performance, with no dependence on $m$. Figure~\ref{fig:narmatanhv} shows the performance for the $\tanh$ transfer function as a function of $v$ on a linear-log scale, clearly showing that for $v<0.01$ the performance of the network equals that of a linear network with, no improvement for $v>0.01$. Figure~\ref{fig:narmataylorv01} shows the performance for $v=0.1$ for various $m$. In this case because of large standard deviation we cannot conclusively say that the increasing nonlinearity in the transfer function is helpful.

\section{Conclusion and outlook}
Nonlinearity of pre-synaptic and dendritic integration plays an important role in the nonlinear computational ability of biological neurons. Similarly, nonlinearity of the transfer function in neural networks is known to increase the capability of the simple multi-layer perceptron to approximate any function. In this work, we systematically studied the effect of increasing nonlinearity on the memory, nonlinear capacity, and the signal-processing performance of echo state networks (ESN), a class of efficient recurrent neural network with state of the art performance in chaotic signal prediction. We found that the region of the $\tanh$ function usually thought of as linear is actually quite nonlinear. Moreover, we found that all the nonlinear power of the $\tanh$ transfer function can be produced using its second-order Taylor approximation. This finding suggests that  ESN performance will benefit from qualitative nonlinearity and not from the degree to which the transfer function is nonlinear. How and why small transfer function nonlinearity helps ESNs will be the subject of our future research.



\section*{Acknowledgment}
This material is $\tanh$ upon work supported by the National Science Foundation under grants CDI-1028238 and CCF-1318833.


\bibliographystyle{IEEEtran}
\bibliography{cisda2015}

\end{document}